# A Novel Two Stream Decision Level Fusion of Vision and Inertial Sensors Data for Automatic Multimodal Human Activity Recognition System

Santosh Kumar Yadav[a,b], Muhtashim Rafiqi[c], Egna Praneeth Gummana[c], Kamlesh Tiwari[c], Hari Mohan Pandey*[d], Shaik Ali Akbar[a,b]

[a]*Academy of Scientific and Innovative Research (AcSIR), Ghaziabad, UP-201002, India*
[b]*Cyber Physical System, CSIR-Central Electronics Engineering Research Institute (CEERI), Pilani-333031, India*
[c]*Department of CSIS, Birla Institute of Technology and Science Pilani, Pilani Campus, Rajasthan-333031, India*
[d]*Department of Computer Science, Edge Hill University, Lancashire, United Kingdom*

**Abstract**

This paper presents a novel multimodal human activity recognition system. It uses a two-stream decision level fusion of vision and inertial sensors. In the first stream, raw RGB frames are passed to a part affinity field-based pose estimation network to detect the keypoints of the user. These keypoints are then pre-processed and inputted in a sliding window fashion to a specially designed convolutional neural network for the spatial feature extraction followed by regularized LSTMs to calculate the temporal features. The outputs of LSTM networks are then inputted to fully connected layers for classification. In the second stream, data obtained from inertial sensors are pre-processed and inputted to regularized LSTMs for the feature extraction followed by fully connected layers for the classification. At this stage, the SoftMax scores of two streams are then fused using the decision level fusion which gives the final prediction. Extensive experiments are conducted to evaluate the performance. Four multimodal standard benchmark datasets (UP-Fall detection, UTD-MHAD, Berkeley-MHAD, and C-MHAD) are used for experimentations. The accuracies obtained by the proposed system are 96.9 %, 97.6 %, 98.7 %, and 95.9 % respectively on the UP-Fall Detection, UTDMHAD, Berkeley-MHAD, and C-MHAD datasets. These results are far superior than the current state-of-the-art methods.

*Email addresses:* santosh.yadav@pilani.bits-pilani.ac.in (Santosh Kumar Yadav), f20180151@pilani.bits-pilani.ac.in (Muhtashim Rafiqi), f20180284@pilani.bits-pilani.ac.in (Egna Praneeth Gummana), kamlesh.tiwari@pilani.bits-pilani.ac.in (Kamlesh Tiwari), profharimohanpandey@gmail.com (Hari Mohan Pandey*), saakbar@ceeri.res.in (Shaik Ali Akbar)

# 1. Introduction

Human activity recognition (HAR) has attracted the attention of many researchers since the 1980s due to its wide range of applications including video surveillance, healthcare, smart-home, sport science, gesture recognition, human-machine interaction, *etc*. [1]. HAR aims to automatically detect and analyze ongoing human activities based on the data obtained from sensors such as RGB cameras, depth cameras, wearable, and inertial sensors. HAR approaches are classified into two groups are (a) vision-based approaches, and (b) sensor-based approaches [2]. The vision-based approach is used to analyze videos or images containing human motions obtained from cameras, while the sensor-based approaches focus on the motion data obtained from smart sensors such as accelerometer, gyroscope, magnetometer, pressure sensors, *etc*. [3].

As a human, we rely on multiple senses such as eyes, nose, ears, tongue, and skin, to navigate the world. In the computer domain, optical sensors are mostly preferred to capture information like human eyes [4]. For HAR, vision-based approaches have been most widely studied [5]. Although it continues to advance, the recognition performance is subject to various challenges such as occlusion, camera position, appearance variations, user's privacy, background clutter, *etc*. [6]. In addition, vision-based approaches are applicable to a limited field of view or a constrained space defined by the camera position and settings [7]. Furthermore, many people feel uncomfortable being continuously monitored by cameras [1]. To avoid the privacy concerns associated with RGB video cameras, researchers have used depth cameras for activity recognition. Depth cameras can provide 3D information, which could be suitable for action recognition. However, they can not be used outdoors because they utilize infrared lights [8]. Moreover, depth cameras are generally not available in common households and are costlier than traditional RGB cameras.

Wearable sensors allow ubiquitous monitoring of activities on a continuous or nearly-continuous basis and are not limited by the constraints of observation space [1]. Therefore, to overcome challenges associated with vision-based approaches, many works of literature have utilized wearable inertial sensors [7]. With the recent advances in the MEMS (microelectromechanical systems) technologies, now wearable inertial sensors are seamlessly integrated with various devices such as smartphones, smartwatches, *etc*., which we generally use in our day-to-day life [9]. However, wearable inertial sensors have limitations related to sensor drift, location of the sensor on the body, and sensor's orientations.

*1.1. Motivation*

As the vision and wearable inertial sensors provide complementary information, using the combination of these two sensing modalities can improve the recognition performance over the individual modality for practical scenarios [12, 10]. Figure 1 demonstrate the stand-to-fall and stand-to-sit activities on the UP-Fall Detection [13] and UTD-MHAD [11] datasets, respectively, using interpolated frames. For the same activity, there is a large difference in data dimensionality, inherent information content, and data distribution between the vision and wearable inertial sensors [14]. Figure 2 presents the data modalities using PGB frames, pose (skeleton), accelerometer, and gyroscope. Simultaneous utilization of vision and wearable inertial sensors can be used for multiple applications such as exergaming, patients monitoring, and assisted living for elderly people [10]. There exist some works, where the fusion of vision and wearable inertial sensors has been incorporated for HAR [15, 16, 17, 18, 19]. Inspired by all these works, in this paper, we focused on multimodal activity recognition using the combination of vision and wearable inertial sensors. However, to avoid the need for specific 3D depth cameras, we choose only RGB videos and wearable inertial sensors as they are readily available and have a lower cost.

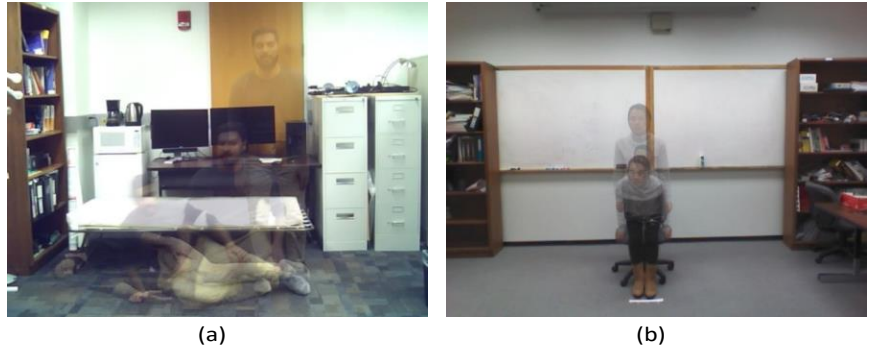

(a)                              (b)

Figure 1: Demonstration of activities using interpolated frames. a) stand-to-fall on the C-MHAD dataset [10], and b. stand-to-sit on the UTD-MHAD [11] dataset.

### 1.2. Challenges and Advancement

Before the resurgence of deep learning methods, traditional approaches for activity recognition used statistical methods to calculate handcrafted features. However, these handcrafted features have limitations associated with the need for expert domain knowledge and calculating exact features, which ultimately limits the model generalization performance on unseen data [20]. In recent years, deep learning based techniques have gained popularity for activity recognition due to their capability of automatic feature extraction. Convolutional neural networks (CNNs) are one of the most widely used deep learning based algorithm for visual applications that automatically learns some representational features [21]. Recurrent neural networks (RNNs) are effective for extracting sequential information from the sensor data. However, they suffer from the challenges of vanishing and exploding gradients [22, 23]. To overcome these challenges and store the information for a longer duration, various RNN architectures are proposed such as long short-term memory (LSTM), gated recurrent unit (GRU), *etc*. [24]. Recently, the combination of CNNs and LSTMs (ConvLSTM) has also been observed for some applications that include face anti-spoofing [25], sentiment analysis [26], and skeletal-based HAR [27].

With the help of recent advances in deep learning methods, now human pose (skeleton) can be estimated from a single image [28, 29] or video [30]. Pose estimated from the RGB images are now more robust compared to skeleton obtained from depth sensors such as Kinect that is efficient for recognizing indoor activities only [31]. Pose data is comparatively lightweight, easier to handle, and very suitable for action recognition than RGB and depth data [32].

### 1.3. Contributions

In this paper, we propose a multimodal HAR system using a combination of vision and inertial sensors. The proposed system has two streams. In the first stream, the video frames obtained from multiple cameras are preprocessed. These preprocessed frames are then passed to a part affinity field network to extract the body joint keypoints. These keypoints are then normalized and inputted into a deep spatiotemporal network. The deep network consists of time-distributed CNNs for spatial feature extraction followed by regularized LSTMs for the temporal features extraction. These spatiotemporal features are then passed to fully connected layers for classification. In the second stream, the raw sensor data obtained from multiple sensing devices are preprocessed. These preprocessed sensor signals are then inputted to specially designed regularized LSTMs for the feature extraction. These features are then passed to fully connected layers for classification. Finally, the Softmax scores of the two streams are fused using decision level fusion. Figure 3 presents the proposed methodology diagram of the proposed system. The system has been tested on four publicly available datasets, namely UP-Fall Detection [13], Berkeley-MHAD [33], UTD-MHAD [11], and C-MHAD [10]. The proposed lightweight system achieves state-of-the-art on these public datasets.

In particular, the major contributions of this paper are as follows.

- We propose a novel two-stream deep spatiotemporal network for multimodal activity recognition using visual and inertial cues. In the first stream, pose keypoints are inputted to specially designed time-distributed CNNs followed by regularized LSTMs for spatial and temporal features extraction. In the second stream, sensor signals obtained from multiple devices are preprocessed. The preprocessed data is then inputted to a novel multi-stream regularized LSTM framework for extracting features from multi-axis sensor measurements such as accelerometers and gyroscopes. Finally, the Softmax scores of vision and inertial streams are fused using the decision level fusion. For fusion, we analyzed average fusion and max fusion. Also, we propose a scheme to handle multiple people present in the background.

- Robustness of the proposed system is tested on the four publicly available benchmark datasets, namely, UP-Fall Detection [13], UTD-MHAD [11], Berkeley-MHAD [33], and C-MHAD [10]. The details of these datasets are described in Section 4.1.

- Extensive computer simulations are conducted on the four multimodal benchmark datasets and the proposed system achieved state-of-the-art on these benchmarks. The accuracies obtained after fusion of visual and inertial cues are 96.9%, 97.6%, 98.7%, and 95.9% on the UP-Fall Detection, UTD-MHAD, Berkeley-MHAD, and C-MHAD datasets, respectively.

- The proposed system includes several other key features over the existing state-of-the-art methods. Key features are: (a) lightweight and simple yet very effective; (b) capable to handle large dataset and include multiple devices and sensors; (c) computational cost-effective, all the experiment can be done using the freely available Google Colaboratory GPU; (d) robust in handling multiple camera videos and data coming from multiple wearable inertial sensors and devices; and (e) very efficient for realistic scenarios.

*1.4. Organization*

The remainder of this paper is organized as follows: Section 2 discusses the related works on multimodal HAR; Section 3 comprehensively presents the proposed methodology; Section 4 elaborates the experimental results on the four publicly available benchmark datasets; and Section 5 presents the concluding remarks and propose future research directions.

## 2. Related Works

This section shade light on the existing work where vision and inertial sensors were used for HAR.

HAR had been developed for a long, however, multimodal HAR is relatively newer. Literature reveals that few researchers have explored multimodal HAR. For example, Delachaux *et al* [16] proposed a multimodal HAR system using the combination of one Kinect camera and five wearable inertial sensors. They utilized a binary trained multi-layer perceptron (MLP) classifier for the classification.

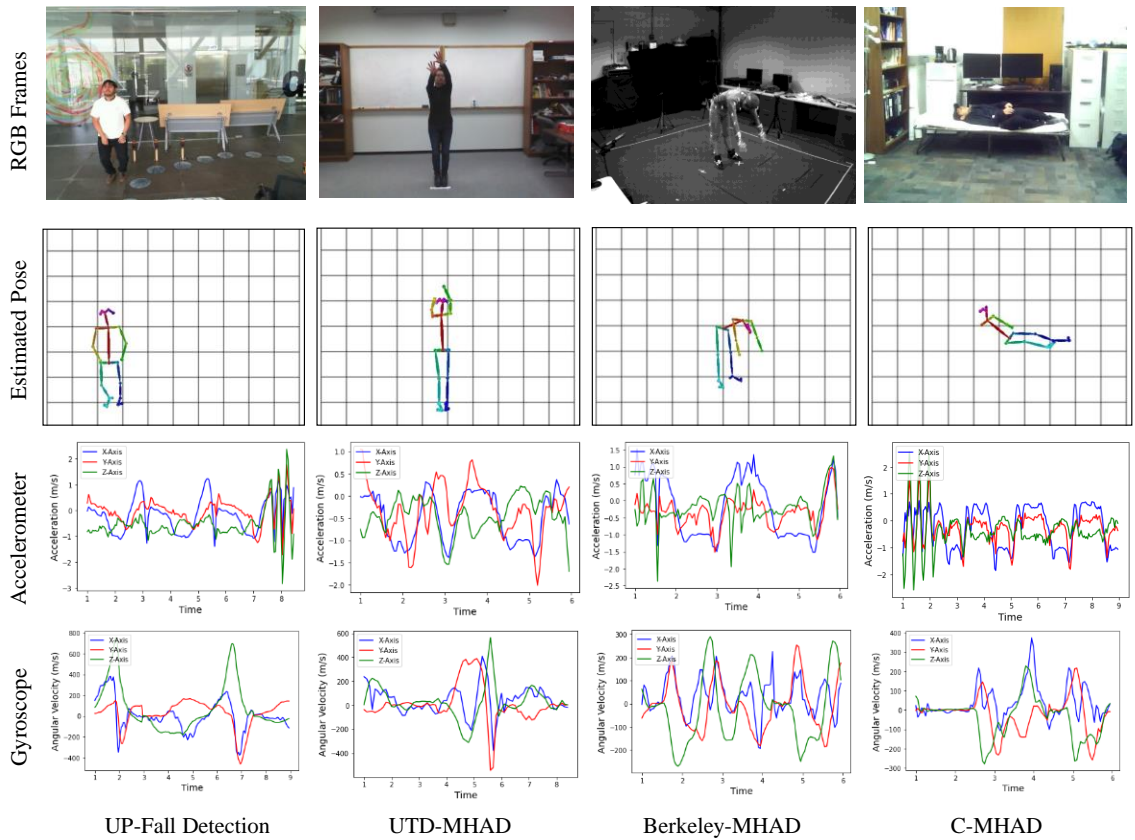

Figure 2: Pictorial representation of different data modalities for the activities - jumping, basketball shoot, bending and stand to lie - from left to right. Row one contains the raw RGB frames from the datasets UP-Fall Detection [13], UTD-MHAD [11], Berkeley-MHAD [33], and C-MHAD [10]. Row two presents the corresponding pose estimated from the raw frames. Row three and four represent the 3-axis accelerometer and gyroscope sensor signals, respectively.

Liu *et al* [18] proposed a multimodal hand gesture recognition system using depth and wearable inertial sensors. The depth sensor provides 3D skeleton joint coordinates while inertial sensors provide 3-axis acceleration and angular velocity signals. They used a hidden Markov model (HMM) framework for the probabilistic classification of five hand gesture classes.

Hondori *et al* [34] fused Kinect and accelerometer sensors data for monitoring the dining activities of post-stroke patients. They calculated 3D trajectories of patient hands and head for monitoring the eating and drinking activities.

Chen *et al* [35] proposed a feature extraction method using the combination of depth and inertial sensors. Kwolek and Kepski [36] proposed a fall detection system using a combination of depth and inertial sensors. They used accelerometers for fall detection while a depth sensor was used for authenticating the fall alerts.

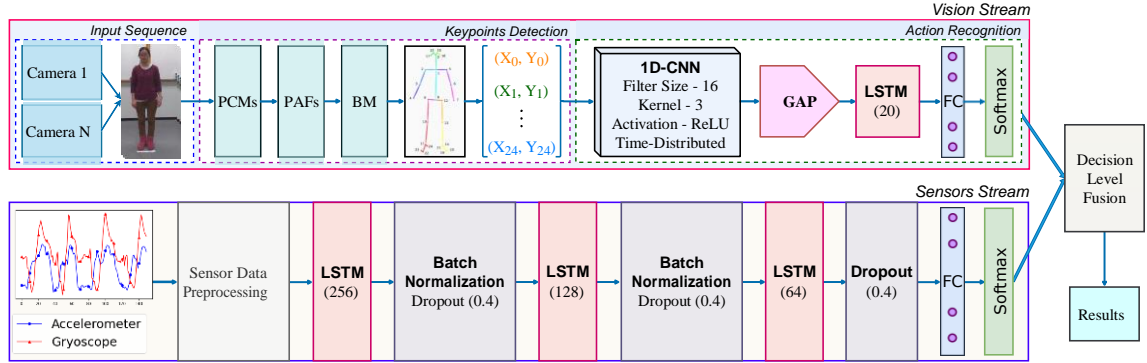

Figure 3: The schematic representation of the proposed system. It has two streams *i.e.* vision stream and inertial stream. In vision stream, frames are acquired from single or multiple cameras. These frames are then passed to pose network, which consist of part confidence maps (PCMs), part affinity fields (PAFs), and bipartite matching (BM), for the keypoints detection. These keypoints are inputted to time distributed 1D-CNN for spatial feature extraction and applied global average pooling (GAP). After that 20 LSTM network is applied for the temporal feature extraction. These spatiotemporal features are passed fully connected layers and calculated the Softmax scores. In the inertial stream, data obtained from accelerometer and gyroscope sensors are preprocessed and inputted to LSTM layers (along with batch normalization and dropout) for feature extraction. These features are passed to fully connected layers and calculated the Softmax scores. Finally, the scores of two streams are fused using decision level fusion.

Gasparrini *et al* [17] proposed a fall detection system using the combination of skeleton joints and accelerometers. Chen *et al* [15] proposed a multimodal activity recognition system using the fusion of depth, skeleton, and inertial sensors data. They used collaborative representation classifiers (CRC) for the classification.

Dawar *et al* [37] proposed an activity recognition system for recognizing the continuous action sequences. They calculated normalized relative orientations (NROs) for skeleton joint coordinates position estimation. Later, they improved it further in [38] to calculate the potential energy functions for skeleton joints using NROs. The acceleration and angular velocity signals were used for removing the false positives. Fuad *et al* [39] first extracted 20 skeleton joint coordinates from a depth sensor camera. Then, skeleton, accelerometer, and gyroscope data are stacked in a column-wise fashion using the bicubic interpolation technique to reduce the temporal variations.

Thanks to the recent advances in deep learning methods, many researchers utilized deep learning approaches for multimodal HAR. Dawar *et al* [40] proposed a deep learning method for action recognition using the combination of depth and inertial sensors to first detect the action and then recognize them from continuous action streams. They used CNNs for depth sensor data and a combination of CNNs and LSTMs for the inertial sensor data. Finally, they combine the decisions of both streams using the decision level fusion.

Ahmad *et al* [20] proposed a multi-stage feature fusion framework for action recognition using the combination of vision and inertial sensors. However, they converted two modalities into multiple modalities that ultimately increases the computational complexity. Additionally, the features had been extracted only from fully connected layers and neglect the important features that could be obtained from CNNs.

Wei *et al* [41] proposed a multimodal activity recognition system using the combination of videos and inertial sensors. They first converted the inertial sensors signals into images, and then applied 2D-CNNs for the feature extraction. Secondly, they applied 3D-CNNs on the video frames to calculate the spatiotemporal features. They explored both the decision level and feature level fusions. Following a similar trend, they further extended their model in [8] and they proposed a continuous activity recognition system. In this, they first detected the actions from continuous streams and then recognized it using 3D-CNNs for videos and 2D-CNNs for inertial sensor images. Finally, they fused the decision of both streams using decision level fusion.

## 3. Proposed Methodology

The proposed system has two synchronized streams *i.e.* inertial stream and vision stream. The methodology of the proposed system is four-fold. First, the vision and inertial sensors data are preprocessed. Second, the preprocessed vision frames are inputted to time-distributed CNN and regularized LSTM networks for spatial and temporal feature extraction, respectively. These spatiotemporal features are then passed to fully connected layers for classification. Third, the preprocessed and cleaned wearable inertial sensor data are inputted to regularized LSTM networks followed by fully connected layers for feature extraction and classification, respectively. Finally, fourth, the Softmax scores obtained from vision and inertial streams are fused using the decision level fusion approach. A schematic representation of the proposed architecture is shown in Figure 3. Further subsections explain the methodology of the proposed system in detail.

### 3.1. Preprocessing

The wearable inertial sensor data obtained from the different devices are time-synchronized before inputting to the network. Since the data captured by the inertial sensors and cameras at a different frequency, data from both the streams are brought down to the same frequency, in order to provide equal weightage to both the streams in the prediction. The raw RGB frames obtained from the vision sensors are inputted to a pose estimation network to calculate the 2D joint locations of the skeleton coordinates. For the pose estimation, a part affinity field network was utilized. It uses the part confidence maps and part affinity fields to locate body joint coordinates followed by bipartite matching for distinguishing the keypoints of multiple individuals. Using this we obtained 25 skeleton joint keypoints of an individual for each frame. Each keypoints have *X* and *Y* coordinates along with their confidence scores. This is further normalized based on the euclidean distance between the keypoints.

The UP-Fall Detection dataset has videos with multiple people in the frame other than the subject of interest. This poses a challenge while detecting the keypoints of interest. To deal with multiple people present in the scene that are unrelated to the task, a suitable method has been proposed. Initially, a window in the center of the frame is used to detect the subject of interest. Once the subject is detected, keypoint co-ordinates from the subsequent frames are selected based on their difference with the previous set of keypoints chosen. This ensures that the keypoints selected are those of the subject irrespective of whether there are multiple people in the frame. All the skeleton joint coordinates, as well as the inertial sensor data obtained from accelerometers and gyroscopes, were normalized using the standard min-max normalization. If *X* represents the training set.

$$X_{scaled} = \frac{X - X_{min}}{X_{max} - X_{min}} \tag{1}$$

A sliding window segmentation technique was used to incorporate the temporal information by introducing temporal order. Different step sizes for the windows were used for the different datasets.

Certain frames were overlapped and carried forward to maintain the dependency in action data. The step sizes and overlap used for UP-Fall Detection, UTD-MHAD, Berkeley-MHAD, and C-MHAD datasets are (50, 30), (50, 10), (50, 10), and (20, 10), respectively. For synchronization of the inertial stream and the keypoints, both streams were brought down to the same frequency. After preprocessing and sliding window segmentation, we obtain sample sizes of 8603, 4841, 10340, and 3790 for each stream for the UP-Fall Detection, UTD-MHAD, Berkeley-MHAD, and C-MHAD datasets, respectively. The train, validation, and test splits were 65%, 10%, and 25%, respectively.

### 3.2. Vision Stream

The preprocessed two-dimensional skeleton joint coordinates are fed to a 16 filter 1D-CNN with a kernel size of 3× 1. CNNs are widely used for analyzing visual imagery. We used 1D-CNNs instead of very complex networks to reduce computational overheads. After feeding the skeleton joints into 1D-CNN, batch normalization and dropout layers (dropout rate of 0.4) are applied for making the model more stable during training and more resistant to overfitting. The output of the CNN layer is flattened and fed to a 20-unit LSTM layer to capture the time-series dependency in the activities. The combination of CNN and LSTM is used to extract spatial and temporal features for activity recognition. Here, spatial features obtained from CNNs are inputted to LSTMs to extract the temporal correlations present in the data. Finally, a Softmax layer is applied to the output of the LSTM layer to give the probabilities for each activity. Figure 3 presents the detailed architecture of the proposed system. A simple yet effective model has been proposed, which is able to capture the patterns in the data and does not compromise on maintaining the time-series dependency.

### 3.3. Inertial Stream

The inertial stream consists of accelerometer, gyroscope, and magnetometer sensor data captured using wearable inertial sensors. Each signal sequence is partitioned into temporal windows. The size of the temporal window varies depending upon the frequency of the sensors employed. The preprocessed and cleaned sensor data is inputted to deep learning networks for feature extraction and classification. Keeping in mind the sequential nature of the data, a pure LSTM based lightweight model was designed for the inertial stream. It allows learning when to forget and update previous hidden states given new information that ultimately helps in maintaining long-term time dependencies. Three LSTM layers with unit sizes 256, 128, and 64 were stacked together. Batch normalization and dropout layers (dropout rate of 0.4) were added in between the LSTM layers for faster convergence and to prevent the model from overfitting on the training set. The output of the last LSTM layer is passed to fully connected layers. Finally, the Softmax scores are calculated. A detailed description of the inertial stream of the proposed system is presented in Figure 3.

### 3.4. Fusion

Decision level fusion was used for the fusion of the Softmax scores of the two streams *i.e.* vision and inertial streams. Since the decision level fusion does not require a separate classifier, it is faster than other fusion techniques [42]. The average fusion (equation 2) and max fusion (equation 3) equations are described as follows.

$$c = \underset{i}{\operatorname{argmax}} \left( \frac{\sum_{j=1}^{n} p(\hat{c}_i | x_j)}{n} \right) \quad (2)$$

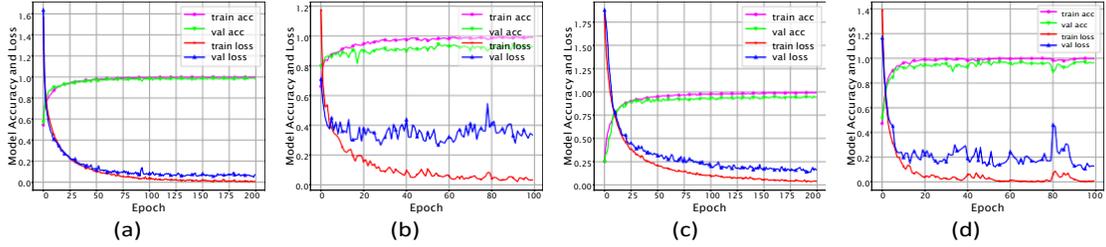

Figure 4: Model accuracy and loss curves for vision and inertial sensors. a. Accuracy and loss for the UP-Fall Detection [13] vision stream, b. Accuracy and loss for the UP-Fall Detection [13] inertial stream, c. Accuracy and loss for the C-MHAD [10] vision stream, and d. Accuracy and loss for the C-MHAD [10] inertial stream.

$$c = \underset{i}{\mathrm{argmax}} \; \underset{j}{\max} \, p(\hat{c}_i | x_j) \qquad (3)$$

where, $c$ denotes the predicted class, $p(\hat{c}_i | x_j)$ is the probability of input $x_i$ belonging to class $c_i$ and n is the number of streams.

For the fusion of classification results of two streams, we used the average fusion. In this, the probability scores of each activity returned by the Softmax layer of each stream were averaged to give the final prediction. The maximum fusion technique was also performed which involved taking the prediction from the stream which was more confident in its prediction. It was found that average fusion outperformed the maximum fusion technique as it takes into account the contribution of both the streams and high confidence false positives are avoided.

## 4. Experimental Results

This section presents the experimental results and it is divided into four subsections. In the first subsection, the dataset details are presented. In the second subsection, the experimental setup is explained which we used to conduct all our experiments. In the third subsection, the model evaluations for different streams are highlighted. The fourth subsection presents the discussions and comparisons of the results with the most recent state-of-the-art methods on the four publicly available benchmark datasets.

### 4.1. Dataset Details

The proposed system has been evaluated on four publicly available datasets, namely, UP-Fall Detection [13], UTD-MHAD [11], Berkeley-MHAD [33], and C-MHAD [10]. In these datasets, we made use of only vision cameras, accelerometers, and gyroscope sensors for our experiments. A brief description of these datasets is given as follows.

### 4.1.1. UP-Fall Detection Dataset

The UP-Fall Detection [13] dataset was published in 2019, with an aim to provide a benchmark to fairly compare activity recognition and fall detection solutions. The UP-Fall Detection dataset was recorded with the help of 17 adult subjects who performed 11 daily activities along with falls, with three attempts each. The daily activities involve six simple activities such as walking, standing, picking up an object, sitting, jumping, and laying. They recorded five types of falls such as falling forward using hands, falling forward using knees, falling backward, falling sitting in an empty chair, and falling sideward. They used multiple modalities namely wearable inertial sensors, ambient sensors, and vision devices. Five wearable sensors were used to collect accelerometer, gyroscope, and ambient light data. In addition to this, an electroencephalograph (EEG) headset, six infrared sensors, and two cameras were used for their dataset.

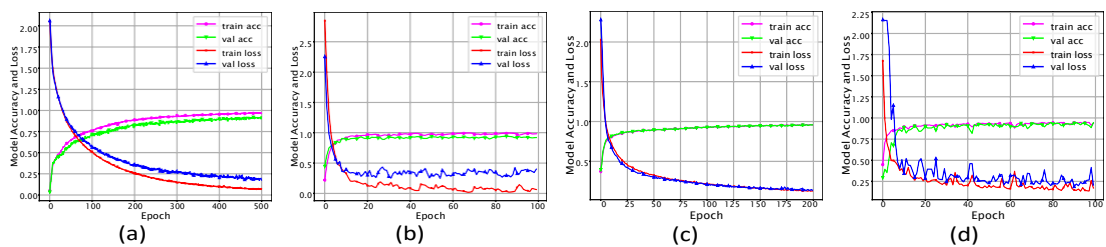

Figure 5: Model accuracy and loss curves for vision and inertial sensors. a. Accuracy and loss for the UTD-MHAD [11] vision stream, b. Accuracy and loss for the UTD-MHAD [11] inertial stream, c. Accuracy and loss for the Berkeley-MHAD [33] vision stream, and d. Accuracy and loss for the Berkeley-MHAD [33] inertial stream.

### 4.1.2. UTD-MHAD Dataset

The UTD-MHAD [11] dataset was published in 2015. It consists of four temporally synchronized data modalities, which include RGB videos, depth videos, skeleton positions, and wearable inertial signals. The dataset was recorded using one wearable inertial sensor and one Microsoft Kinect camera. It contains 27 actions performed by 8 subjects (4 females and 4 males), where each action is repeated 4 times by all the subjects.

### 4.1.3. Berkeley-MHAD Dataset

The Berkeley-MHAD [33] dataset was proposed in 2013. It consists of temporally synchronized data from an optical motion capture system, multi-baseline stereo cameras from multiple views, depth sensors, accelerometers, and microphones. The dataset contains 11 actions performed by 7 male and 5 female subjects with each action being performed 5 times by each subject. Each action was simultaneously captured by five different systems: an optical motion capture system, four multi-view stereo vision camera arrays, two Microsoft Kinect cameras, six wireless accelerometers, and four microphones. Multi-view video data was captured by 12 cameras which were arranged into four clusters: two clusters for stereo and two clusters with four cameras for multi-view capture. Six three-axis wireless accelerometers were used to measure movement at the wrists, ankles, and hips.

### 4.1.4. C-MHAD Dataset

The Continuous Multimodal Human Action Dataset (C-MHAD) [10] was released in 2020. It is a first of its kind multimodal activity recognition dataset in which video and inertial data stream were captured simultaneously in a continuous way. The inertial signals consist of three-axis acceleration signals and three-axis angular velocity signals captured by a wearable inertial sensor at a frequency of 50 Hz. The videos are captured by a laptop video-cam at a rate of 15 image frames per second. The dataset contains two sets of actions corresponding to smart TV gestures and transition movements. The continuous dataset for the transition movements contains seven actions performed by 12 subjects (ten males and two females). For each subject, ten continuous streams of video and inertial data, each lasting for 2 min, were captured.

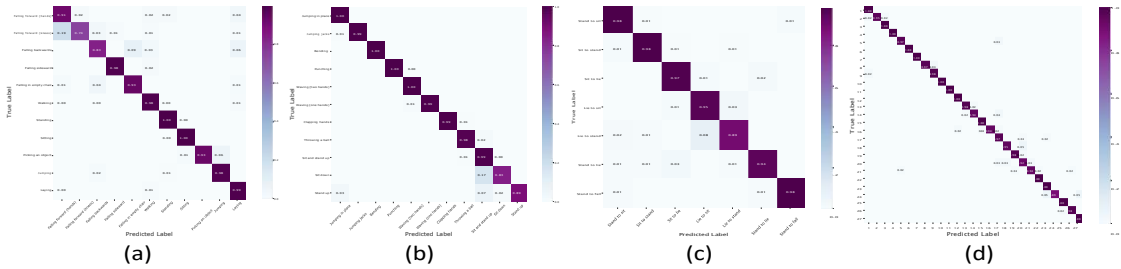

(a) (b) (c) (d)

Figure 6: Confusion matrices after fusing the vision and inertial streams on the a. UP-Fall Detection [13], b. Berkeley-MHAD [33] dataset, c. C-MHAD [10], and d. UTD-MHAD [11] datasets.

Table 1: Results of the proposed model on the UP-Fall Detection [13], UTD-MHAD [11], Berkeley-MHAD [33], and C-MHAD [10] datasets.

| Dataset | Model Name | Accuracy | Precision | Recall | F1-Score |
|---|---|---|---|---|---|
| UP-Fall Detection | Inertial Stream | 93.3% | 93.4% | 93.3% | 93.2% |
|  | Vision Stream | 96.7% | 96.9% | 96.7% | 96.6% |
|  | Fusion | 96.9% | 97.1% | 96.9% | 96.9% |
| UTD-MHAD | Inertial Stream | 94.7% | 95% | 94.8% | 94.9% |
|  | Vision Stream | 95.8% | 95.8% | 95.8% | 95.8% |
|  | Fusion | 97.6% | 97.7% | 97.6% | 97.6% |
| Berkeley-MHAD | Inertial Stream | 95.6% | 95.4% | 95.6% | 95.3% |
|  | Vision Stream | 98.7% | 98.7% | 98.7% | 98.7% |
|  | Fusion | 98.7% | 98.7% | 98.7% | 98.7% |
| C-MHAD | Inertial Stream | 93.5% | 93.5% | 93.5% | 93.4% |
|  | Vision Stream | 95.0% | 95.1% | 95.0% | 95.0% |
|  | Fusion | 95.9% | 95.9% | 95.9% | 95.9% |

*4.2. Experimental Settings*

All the experiments were conducted using Google Colaboratory, making the use of the freely available GPU it provides. This also ensures that the proposed system is effective and can run with limited resources. The codes were written in Python and made use of the Keras framework with TensorFlow backend, for building the models. The proposed system was trained using a batch size of 32 after repeated hyper-parameter tuning. For the inertial stream, the LSTM layer used a unit forget bias of 0.5, and all the experiments were run for 100 epochs. For the vision stream, all the experiments were run for 200 epochs except on the UTD-MHAD dataset. For the UTD-MHAD vision stream, it took 500 epochs to converge given a large number of activities present compared to other datasets. Both the streams made use of the Adam optimizer with a learning rate of 0.0001 on the categorical cross-entropy loss function. Figure 4 and Figure 5 show the accuracy and loss curves for the vision and inertial streams on the four datasets. The curves validate the effectiveness of the proposed model and demonstrate that it is a good fit for both the training and validation sets.

Table 2: Performance comparison of our proposed model with other state-of-the-art models on the Berkeley-MHAD [33] dataset. The table is sorted based on the modality. There were no works in which skeleton and inertial sensors combination were used on this dataset, so a direct comparison is difficult. D-Depth, I-Inertial, S-Skeleton, V-Vision. **Red** denotes the best results for the modality we used. **Blue** denotes the best results.

| Authors | Year | Method | Modality | Accuracy |
|---|---|---|---|---|
| Shafaei *et al.* [43] | 2016 | Dense Depth Classification | D | 98.1% |
| Ofli *et al.* [33] | 2013 | Multiple Kernel Learning | D, I | 97.81% |
| Chen *et al.* [35] | 2014 | Depth Motion Maps and CRC | D, I | 99.13% |
| Chen *et al.* [35] | 2014 | Depth Motion Maps and Sparse Representation Classifiers | D, I | 99.54% |
| Ahmad *et al.* [20] | 2019 | Deep Hybrid Fusion Framework | D, I | 99.8% |
| Ahmad *et al.* [44] | 2020 | Multistage Gated Average Fusion | D, I | **99.85%** |
| Ofli *et al.* [33] | 2013 | Multiple Kernel Learning | D, V | 99.27% |
| Liu *et al.* [14] | 2020 | Semantics-aware Adaptive Knowledge Distillation Networks | V, I | 99.33% |
| **Proposed Method** | **2020** | **Proposed** | **S, I** | **98.7%** |

Table 3: Performance comparison of our proposed model with other state-of-the-art models on the UTD-MHAD [11] dataset. The table is sorted based on the modality. D-Depth, I-Inertial, S-Skeleton, V-Vision. **Red** denotes the best results for the modality we used. **Blue** denotes the best results.

| Authors | Year | Method | Modality | Accuracy |
|---|---|---|---|---|
| Lemieux *et al.* [9] | 2020 | 1D-CNN based architecture | I | 85.35% |
| Liu *et al.* [31] | 2018 | Heatmap | V | 92.84% |
| Memmesheimer *et al.* [45] | 2020 | Gimme Signals | S | 93.33% |
| Liu *et al.* [31] | 2018 | Heatmaps and Pose | D | 94.51% |
| Chen *et al.* [11] | 2015 | Depth Motion Maps and CRC | D, I | 79.1% |
| Dawar *et al.* [46] | 2018 | Weighted Depth Maps (CNN) and Inertial (CNN + LSTM) | D, I | 89.2% |
| Ehatisham *et al.* [12] | 2019 | K-NN | D, I | 97.6% |
| Ahmad *et al.* [20] | 2019 | Deep Hybrid Fusion Framework | D, I | 99.3% |
| Ahmad *et al.* [44] | 2020 | Multistage Gated Average Fusion | D, I | **99.3%** |
| Madany *et al.* [47] | 2016 | Multiview Discriminative Analysis of Canonical Correlations | D, I, S | 93.26% |
| Wei *et al.* [41] | 2019 | RGB (3D-CNN) and Inertial(2D-CNN) | V, I | 95.6% |
| Liu *et al.* [14] | 2020 | Semantics-aware Adaptive Knowledge Distillation Networks | V, I | 98.60% |
| Zhu *et al.* [32] | 2018 | RGB (I3D) and Skeleton (I3D + 2D-Deconv + 2D-CNN) | V, S | 92.5% |
| Verma *et al.* [48] | 2020 | Skeleton (CNN + LSTM) + RGB (CNN) | V, S | 96.5% |
| Imran *et al.* [19] | 2020 | RGB (2D-CNN) + I (1D-CNN) + S (RNN) | V, I, S | 97.91% |
| Chen *et al.* [49] | 2015 | Depth Motion Maps and CRC | S, I | 91.5% |
| Fuad *et al.* [39] | 2018 | Single layer Neural Network | S, I | 95% |
| **Proposed Method** | **2020** | **Proposed** | **S, I** | **97.6%** |

*4.3. Performance Evaluation*

Four datasets were used to evaluate the proposed system. The evaluation metrics used are accuracy, precision, recall, and F1-Score. The precision, recall, and F1-Score are calculated using the macro-avg for different activities. The results on individual streams are presented in Table 1, to make a comparison with the final fused results. The following sections present the results of the evaluation on each dataset.

Table 4: Performance comparison of our proposed model with the state-of-the-art on the C-MHAD [10] dataset for transition movements. I-Inertial, S-Skeleton, V-Vision. **Red** denotes the best result.

| Authors | Year | Method | Modality | Precision | Recall | F1-Score | Accuracy |
|---|---|---|---|---|---|---|---|
| Wei *et al.* [10] | 2020 | CNN | V, I | - | - | 78.8% | - |
| **Proposed Method** | **2020** | **Proposed** | **S, I** | **95.9%** | **95.9%** | **95.9%** | **95.9%** |

*4.3.1. Results on the UP-Fall Dataset*

Table 1 shows the results obtained using the proposed model on the UP-Fall Detection dataset. The inertial and vision streams give an accuracy of 93.3% and 96.7%, respectively. The fusion of the two streams gives a higher accuracy of 96.9% compared to individual streams. Figures 4(a) and 4(b) shows the accuracy and loss plots for the model on this dataset, respectively. It is evident from the curve that the vision model starts converging towards the end of 50 epochs while the inertial stream starts converging after being trained for 40 epochs. The confusion matrix for class-wise accuracy is given in Figure 6. From the confusion matrix, it can be inferred that except for 'falling forward using knees' and 'falling backwards', the accuracy for all the other activities is higher than 90%. 'Falling forward using hands' and 'falling forward using knees' are activities of a similar kind, and hence there is a higher misclassification error of 26% for this activity.

*4.3.2. Results on the UTD-MHAD Dataset*

Table 1 shows the results obtained on the UTD-MHAD Dataset. The inertial stream of the proposed system scored an accuracy of 94.7%, while the vision stream scored an accuracy of 95.8%. After the fusion of the two streams, 97.6% accuracy is obtained. Figures 5(a) and 5(b) shows the accuracy and loss curves for the proposed model on this dataset, respectively. The curves reveal that the inertial stream converges faster, after training for around 20 epochs. For the vision stream, the model takes longer to converge. Initially, the model for vision stream was trained for 200 epochs, but since the model had the potential to learn, it was further trained for 500 epochs. The confusion matrix for class-wise accuracy is given in Figure 6(d). The confusion matrix shows that 13 out of the 27 activities scored 100% classification accuracy. The minimum classification accuracy is 89% for the 'sit-to-stand' activity.

*4.3.3. Results on the Berkeley-MHAD Dataset*

Table 1 shows the results obtained on the Berkeley-MHAD dataset. The proposed model achieves an accuracy of 95.6% and 98.7% on the inertial and vision stream, respectively. The accuracy after fusion of inertial and vision stream is 98.7%. The accuracy and loss curves in Figure 5(c) and 5(d) demonstrate that the model is a perfect fit on the dataset, respectively. The confusion matrix for class-wise accuracy given in Figure 6(b), shows that there is relatively a larger error in classifying the activity 'sit down' and activity 'sit down and stand up'. As the two activities are closely related and overlapping, the features learned for both the activities by our model are similar which explains the relatively higher error on this activity compared to the other 9 activities. The higher error is also attributed to the fact that the relative abundance of the two activities in the dataset is not the same.

*4.3.4. Results on the C-MHAD Dataset*

Opposed to other datasets, the C-MHAD dataset, as the name suggests, is a continuous multimodal HAR dataset. While using this dataset, we segmented the actions of interest for the purpose of evaluating our model instead of using it as a continuous dataset. Table 1 shows the results obtained on the C-MHAD dataset. An accuracy of 93.5% and 95.0% is obtained using the inertial and vision streams, while fusion gives a higher accuracy of 95.9%. Figure 4(c) and 4(d) shows the accuracy and loss plots for the model on this dataset, respectively. The confusion matrix for class-wise accuracy is given in Figure 6(c). It shows that even though the activities performed are closely related and overlapping, the model performs well as 5 out of 7 activities score an accuracy of above 95%.

Table 5: Performance comparison of our proposed model with other state-of-the-art models on the UP-Fall Detection [13] dataset. The table is sorted based on the modality. I-Inertial, S-Skeleton, V-Vision. **Red** denotes the best results.

| Authors | Year | Method | Modality | Precision | Recall | F1-Score | Accuracy |
|---|---|---|---|---|---|---|---|
| Martínez *et al.* [13] | 2019 | kNN | V | 68.82% | 58.49% | 60.51% | 92.06% |
| Martínez *et al.* [13] | 2019 | Random Forest | V | 75.52% | 66.23% | 69.36% | 95.09% |
| Martínez *et al.* [13] | 2019 | CNN | V | 71.8% | 71.3% | 71.2% | 95.1% |
| Espinosa *et al.* [50] | 2020 | KNN | V | 16.32% | 14.35% | 15.27% | 27.30% |
| Espinosa *et al.* [50] | 2020 | Random Forest | V | 14.45% | 14.30% | 14.37% | 29.30% |
| Espinosa *et al.* [50] | 2020 | MLP | V | 9.05% | 11.03% | 9.94% | 30.08% |
| Espinosa *et al.* [50] | 2020 | SVM | V | 14.03% | 14.10% | 14.06% | 32.40% |
| Espinosa *et al.* [51] | 2019 | CNN | V | 74.25% | 71.67% | 72.94% | 82.26% |
| Kraft *et al.* [52] | 2020 | CNN | I | 95.2% | 94.0% | 94.2% | - |
| Chahyati *et al.* [53] | 2020 | Majority Voting (CNN + LSTM) | V, I | 95.64% | 95.29% | 95.44% | - |
| Martínez *et al.* [13] | 2019 | MLP | V, I, EEG | 77.7% | 69.9% | 72.8% | 95.0% |
| **Proposed Method** | **2020** | **Proposed** | **S, I** | **96.9%** | **97.1%** | **96.9%** | **96.9%** |

*4.4. Discussion and Comparison*

We compare the proposed system's performance with other state-of-the-art activity recognition works on the four public datasets reported in the literature. The comparison results on these four datasets *i.e.* UP-Fall Detection [13], UTD-MHAD [11], Berkeley-MHAD [33], and C-MHAD [10] are presented in Table 5, 3, 2, and 4, respectively. For a fair comparison, we mentioned the modalities used by each paper along with their methods. The tables are sorted based on their modality and results. The proposed system utilizes inertial sensors and skeleton keypoints obtained from pose estimation for model implementations and predictions.

The performance comparison on the UP-Fall Detection dataset is given in Table 5. Our model was compared with [13, 51, 50, 52, 53, 54]. Chahyati *etal.* [53] have utilized the majority voting fusion of vision and inertial streams. The average level decision fusion of our proposed model outperforms the majority voting method since averaging gives equal weightage to information from both the streams, unlike majority voting where the prediction depends on the number of frames/sequences available for each stream. The proposed system achieves state-of-the-art performance on the UP-Fall Detection dataset even when compared with other works using additional inertial sensors, and also depth sensors.

The performance comparison on the UTD-MHAD dataset and the Berkeley-MHAD dataset is presented in Table 3 and Table 2, respectively. For the UTD-MHAD dataset, our model was compared with [49, 11, 47, 31, 46, 39, 32, 20, 12, 41, 48, 45, 19, 9, 14, 44]. A direct comparison of our proposed system on the Berkeley-MHAD dataset is not possible because no other paper used the skeleton and inertial data together on this dataset. However, we presented the state-of-the-art results of other modalities [33, 43, 35, 20, 14, 44]. Ahmad *etal.* [44] used depth and inertial sensors for testing their model performance on the UTD-MHAD and Berkeley-MHAD datasets. In comparison with the state-of-the-art results on these two datasets by [44], the results of the proposed system are competitive, even though it makes use of only inertial sensors and pose estimated from RGB frames. Propose system gives a significant improvement on the work by [39] on the UTD-MHAD dataset, which also attempts to use sensor and skeleton information for the classification. It is to be noted that the proposed system does not require any special hardware such as depth sensors or 3D cameras.

The proposed system is also evaluated on the relatively newer C-MHAD dataset, which was proposed by Wei *et al.* [10] in 2020. The creators of the dataset have evaluated the continuous dataset on action detection along with recognition. C-MHAD is used as a segmented dataset in our experiment and the proposed system has obtained satisfactory results. The proposed system achieves an overall 95.9% accuracy on the C-MHAD dataset. Table 4 presents the comparison results on the C-MHAD dataset. Since this dataset is relatively newer and first of its kind, there are no other benchmark models to compare our results except the base results provided by [10].

Overall, the proposed system achieved state-of-the-art results on these four datasets using the inertial signals and pose estimated from the RGB frames. It is to be noted that the proposed model is lightweight and all the experiments can be performed using the open-source Google Colaboratory for such complex datasets.

## 5. Conclusion

This paper presented a multimodal HAR system using the fusion of vision and inertial sensors. The proposed system utilizes two streams. The first stream is for vision modality, and the second stream is for the inertial sensors modality. Vision stream utilizes a pose estimation based approach to detect the user's keypoints. These keypoints are then preprocessed and inputted to time-distributed CNN networks in a sliding window fashion for spatial feature extraction followed by regularized LSTM networks to obtain the temporal features. The outputs of the regularized LSTM networks are inputted to fully connected layers for classification. In the second stream, inertial sensors data are preprocessed and inputted to specially designed LSTM networks for feature extraction. These features are then passed to fully connected layers for the classification. Finally, the Softmax scores of the vision and inertial streams are fused using the decision level fusion. The proposed model has been tested on four benchmark datasets *i.e.* UP-Fall Detection [13], UTD-MHAD [11], Berkeley-MHAD [33], and C-MHAD [10] datasets. The proposed system achieves state-of-the-art on these datasets for inertial and skeleton modalities in terms of both recognition accuracy and computational cost. The proposed lightweight and simple model has been proven to be very effective.

## Acknowledgements

The authors would like to thank anonymous reviewers and our parent organizations for extending their support for the betterment of the manuscript. We appreciate the assistance provided by CSIR, India.

## Conflict of Interest

The authors declare that they have no conflict of interest.